\documentclass[10pt,twocolumn,letterpaper]{article}

\usepackage{cvpr}              

\usepackage[dvipsnames]{xcolor}

\definecolor{cvprblue}{rgb}{0.21,0.49,0.74}

\usepackage[pagebackref,breaklinks,colorlinks,citecolor=cvprblue]{hyperref}
\usepackage{xcolor,colortbl}
\usepackage{bbm}
\usepackage[accsupp]{axessibility}

\newcommand{\tp}{\textsuperscript}

\definecolor{Gray}{gray}{0.85}
\definecolor{BA}{rgb}{0.98, 0.91, 0.71}

\def\paperID{15} 
\def\confName{CVPR}
\def\confYear{2024}

\title{MoDA: Leveraging Motion Priors from Videos for \\Advancing Unsupervised Domain Adaptation in Semantic Segmentation}

\author{
\hspace{0mm}Fei Pan$^{1}$
\hspace{5mm}Xu Yin$^{2}$
\hspace{5mm}Seokju Lee$^{3}$
\hspace{5mm}Axi Niu$^{2}$
\hspace{5mm}Sungeui Yoon$^{2}$
\hspace{5mm}In So Kweon$^{2}$
\vspace{1mm}
\\
$^{1}$University of Michigan, Ann Arbor
\hspace{5mm}
$^{2}$KAIST
\hspace{5mm}
$^{3}$KENTECH
\\
\hspace{0mm}{\tt\small 
feipan@umich.edu, slee@kentech.ac.kr, \{yinofsgvr,sungeui,iskweon77\}@kaist.ac.kr}
}

\begin{document}

\maketitle

\begin{abstract}
Unsupervised domain adaptation (UDA) has been a potent technique to handle the lack of annotations in the target domain, particularly in semantic segmentation task. This study introduces a different UDA scenarios where the target domain contains unlabeled video frames. Drawing upon recent advancements of self-supervised learning of the object motion from unlabeled videos with geometric constraint, we design a \textbf{Mo}tion-guided \textbf{D}omain \textbf{A}daptive semantic segmentation framework (MoDA). MoDA harnesses the self-supervised object motion cues to facilitate cross-domain alignment for segmentation task. First, we present an object discovery module to localize and segment target moving objects using object motion information. Then, we propose a semantic mining module that takes the object masks to refine the pseudo labels in the target domain. Subsequently, these high-quality pseudo labels are used in the self-training loop to bridge the cross-domain gap. On domain adaptive video and image segmentation experiments, MoDA shows the effectiveness utilizing object motion as guidance for domain alignment compared with optical flow information. Moreover, MoDA exhibits versatility as it can complement existing state-of-the-art UDA approaches. Code:
\href{https://github.com/feipanir/MoDA}{https://github.com/feipanir/MoDA}.

\end{abstract}

\section{Introduction}
\begin{figure}[t]
    \centering
    \includegraphics[width=0.5\textwidth]{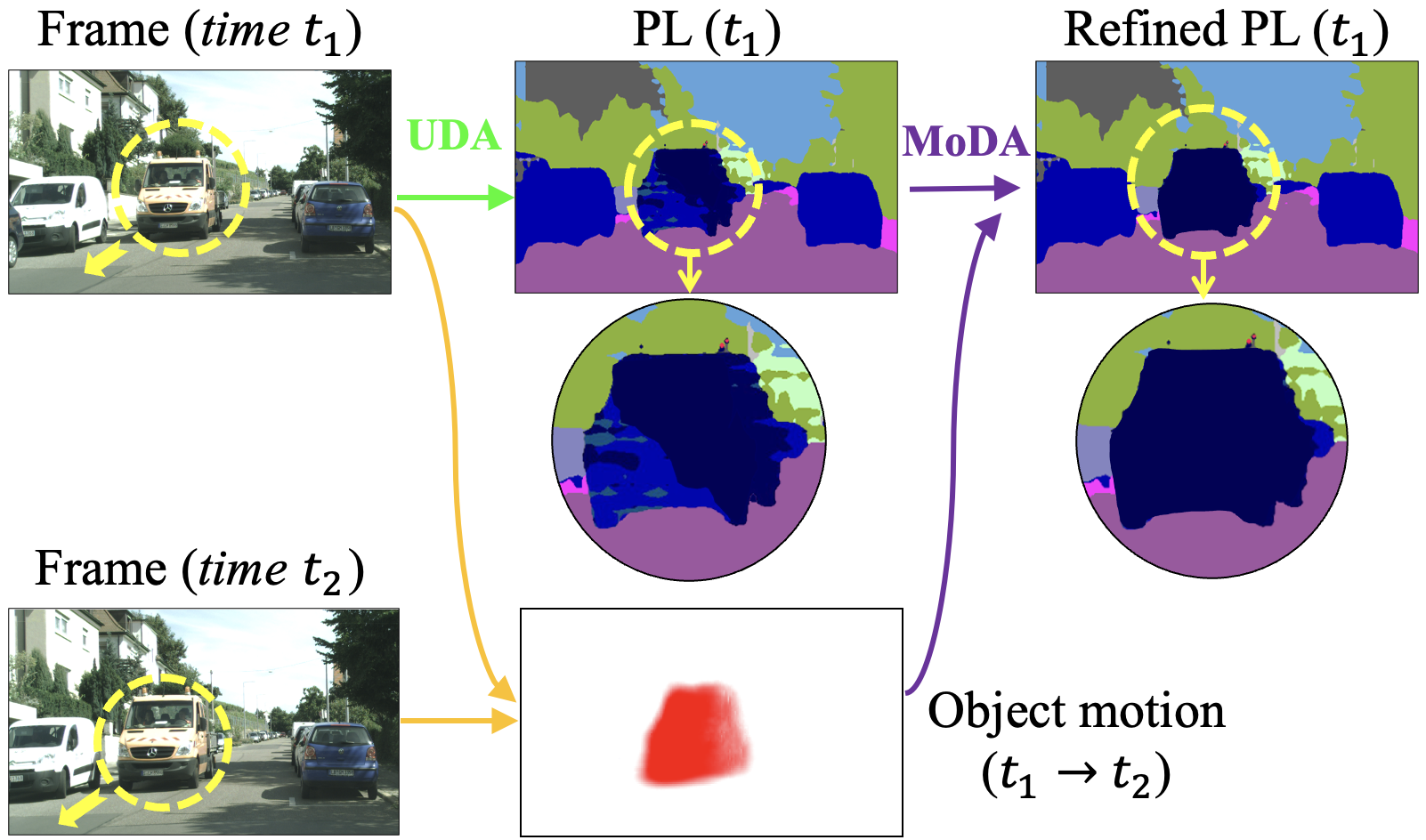}
    \caption{Current \textcolor{green}{UDA} methods show notable performance for background categories (\textit{e.g.}, tree, road, sky), but they are limited to the real-world dynamic scenes containing multiple moving objects (\textit{e.g.}, buses). We propose \textcolor{violet}{MoDA} that uses object motion from unlabeled videos as complementary guidance to refine pseudo labels (PL) in the target domain. Note that the object motion is learned using using \textit{ self-supervised geometric constraints} from sequential video frames ($t_1$ and $t_2$), \textit{without requiring any annotations}.}
    \label{fig:teaser}
\end{figure}

Fully-supervised semantic segmentation~\cite{FCN,deeplabv2} is a data-hungry task that requires all pixels of training images to be assigned with a semantic label. However, providing pixel-wise human annotations for semantic segmentation is expensive and time-consuming~\cite{cityscapes}. Recently, unsupervised domain adaptation (UDA) has become an effective technique to alleviate the necessity of pixel-wise data labeling. Existing UDA for semantic segmentation methods utilizes adversarial learning for feature or output-level domain alignment~\cite{advent, adaptsegnet, IntraDa} or self-training techniques that refine target pseudo labels in an iterative process~\cite{da_sac, CBST, DACS, CRST}. While these methods show notable improvements, particularly for background categories (\textit{e.g.}, tree, road, sky), they are limited to the real-world dynamic scenes containing multiple moving objects, as an example shown in Fig.~\ref{fig:teaser}.

The recent trend in dynamic scene understanding involves learning the object motion and the camera's ego-motion from unlabeled sequential image pairs. A bird's eye view in Fig. \ref{fig:geometric_training}(a) shows an example of the \textit{object motion} and the \textit{camera's ego-motion}. Existing works~\cite{motion_net, 3dmotion_seokju, 3dmotion_2} suggest learning a motion network to separate the local object motion from the global camera ego-motion in the static scenes with \textit{self-supervised geometrical constraints}. This separation helps to isolate the object motion from the camera's ego-motion. As an example in Fig.~\ref{fig:teaser}, the object motion is capable of \textit{segmenting} the moving yellow bus out from static scenes. 

We propose that the object motion learned from geometric constraints can be used as \textit{complementary} guidance for domain adaptation to the target domain. Specifically, the segmentation network suffers from the cross-domain gap due to the lack of semantic labels in the target domain. However, the motion network is trained separately from the segmentation network. Moreover, the motion network is trained only using target frames in the \textit{self-supervised} manner with geometric constraints, which \textit{does not require} any semantic labels. So the object motion from the motion network is not affected by the cross-domain gaps caused by the labeling issue.

In this work, we present a motion-guided unsupervised domain adaption (MoDA) method for segmentation task, which utilizes the \textit{self-supervised object motion from geometrical constraints as cues for domain alignment}. First, we present an object discovery module that takes the object motion information learned from the unlabeled videos as input and localizes and segments the moving object in the target domain. Then, we propose a semantic mining module to refine target pseudo labels by utilizing the object masks from the target domain. Subsequently, these high-quality target pseudo labels are sent as input in the self-training loop to optimize the target segmentation model. On domain adaptive video and image segmentation benchmarks,  MoDA shows the effectiveness utilizing object motion as guidance for domain alignment compared with optical flow information. Moreover, it is versatile to complement existing state-of-the-art UDA approaches.

\noindent \textbf{Contribution:}
\begin{itemize}

\item{We are the first to utilize self-supervised object motion information from unlabeled videos to facilitate cross-domain alignment for semantic segmentation task, without requiring any target annotations.}

\item{MoDA contains the object discovery module and the semantic mining module to refine target pseudo labels. These two modules are directly used without the need for training.}

\item{MoDA shows superior performance on the benchmarks for the domain adaptive video and image segmentation compared with optical flow baselines. MoDA is also versatile to complement existing state-of-the-art UDA approaches.}

\end{itemize}

\section{Related Works}
\noindent\textbf{Domain adaptive image segmentation.}
The goal of domain adaptation is to align the domain shift between the labeled source and target domains~\cite{electronics3,electronics4}. For domain adaptive image segmentation, adversarial learning~\cite{advent, adaptsegnet, IntraDa, li2020generating, wang2019weakly} and self-training~\cite{ProDA, SePiCo, CRST, CBST, DACS,da_sac, rectifying,LTIR, lu2023uncertainty, fp_2, fp_3} approaches are widely adopted, and demonstrate compelling adaption results. \cite{IntraDa} designs two discriminator networks to implement the inter-domain and intra-domain alignment. \cite{MRNet} proposes to average the predictions from the source and the target domain to stabilize the self-training process and further incorporate the uncertainty~\cite{rectifying} to minimize the prediction variance. 
Existing works consider the domain alignment on the category level~\cite{wang2022cluster} and instance level~\cite{yuan2022birds} to learn domain-invariant features. ~\cite{zhao2022source} proposes to handle more diverse data in the target domain. adopt image-level annotations from the target domain to bridge the domain gap. ~\cite{GUDA} propose combined learning of depth and segmentation for domain adaptation with self-supervision from geometric constraints. Different from existing approaches, we consider using motion priors as guidance for domain alignment in this work. 

\noindent\textbf{Self-training for domain alignment.} In self-training, the network is trained with pseudo-labels from the target domain, which can be pre-computed offline or calculated online during training~\cite{ProDA, SePiCo, da_sac, daformer, HRDA}. ~\cite{ProDA} proposes to estimate category-level prototypes on the fly and refine the pseudo labels iteratively, to enhance the adaptation effect.~\cite{da_sac} utilizes data augmentation and momentum updates to regulate cross-domain consistency. ~\cite{zhou2020affinity} proposes to align the cross-domain gaps on structural affinity space for the segmentation task.
In this work, we introduce the motion masks that provide complementary object geometric information as prior. Specifically, we develop the motion-guided self-training and the moving object label mining module to refine the target pseudo labels and thus improve the adaptation performance.

\noindent\textbf{Domain adaptive video segmentation.}  Exploiting motion information like optical flow~\cite{optical_flow_da} to separate the objects in videos to regulate the segmentation training is well explored in the video segmentation field. In UDA, there are several attempts that introduce temporal supervision signals to enforce the domain alignment.~\cite{DA-VSN} regulates the cross-domain temporal consistency with adversarial training to minimize the distribution discrepancy.~\cite{TPS} proposes to capture the spatiotemporal consistency in the source domain by data augmentation across frames. In this work, we propose to utilize the 3D object motion ~\cite{motion_net} of the target sequential image pairs, which provides rich information for localizing and segmenting the moving objects. 
\section{Preliminary}
\begin{figure*}[t!]
    \centering\includegraphics[width=0.89\textwidth]{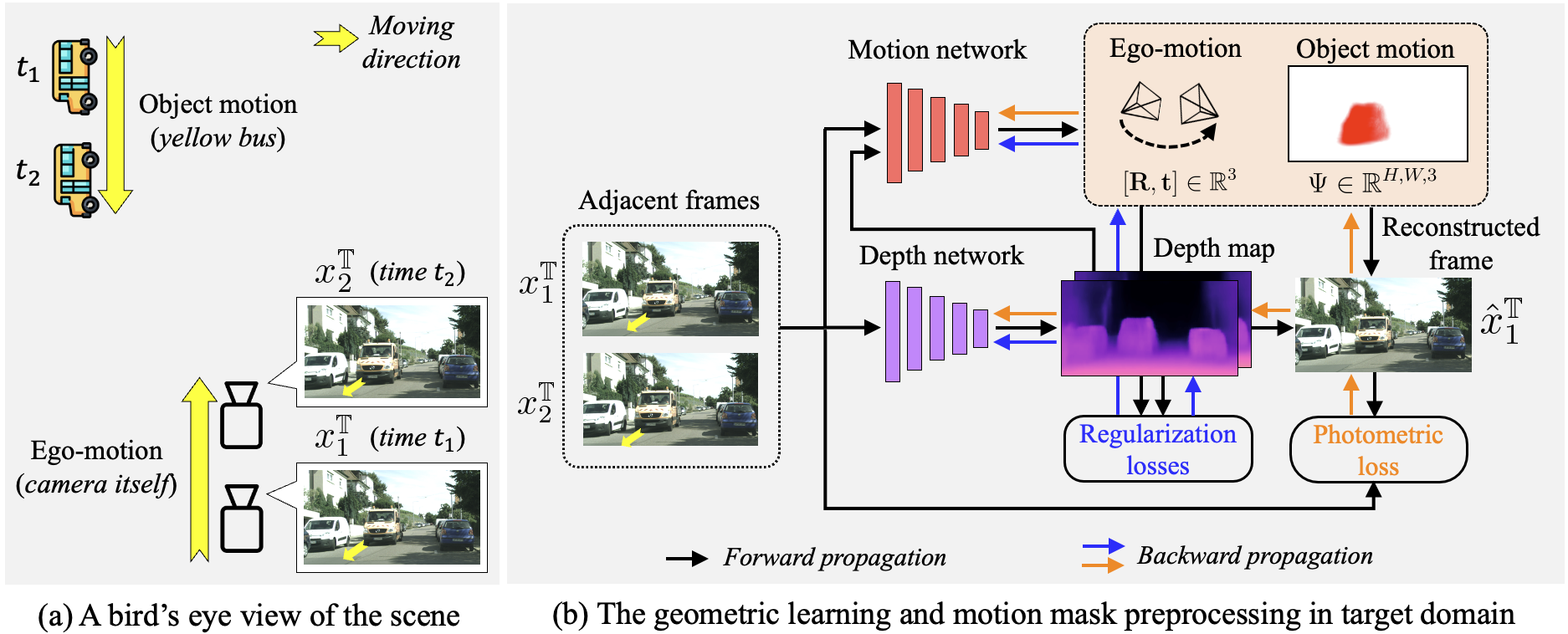}
    \caption{The object motion information is learned by self-supervised geometric constraints from unlabeled target video frames, without any annotations. (a) The visualization of a bird's eye view of the dynamic scene, where the yellow bus is moving toward the camera. We indicate the \textit{object motion} of the yellow bus and the \textit{ego motion} from the camera itself. (b) The diagram for geometric training to learn the object motion from a pair of target adjacent video frames. The motion network and depth network are trained by the \textit{self-supervised} losses (\textcolor{orange}{photometric loss} and \textcolor{blue}{regularization losses}) following geometric constraints.}
    \label{fig:geometric_training}
\end{figure*}
We train the motion network $G_{m}$ to learn object motion prediction in the target domain illustrated in Sec.~\ref{geometric_learning}. Then, we conduct motion mask preprocessing to obtain the instance-level motion masks in the target domain shown in Sec.~\ref{motion mask preprocessing}. 

\noindent \textbf{Notations.} We have a set of source images $X^{\mathbb{S}}=\{x^{\mathbb{S}}_n\}_{n=1}^{N^{\mathbb{S}}}$ with 
the corresponding segmentation annotations $Y^{\mathbb{S}}=\{y^{\mathbb{S}}\}_{n=1}^{N^\mathbb{S}}$, where $N^{\mathbb{S}}$ is the number of the source images. We also have a set of unlabeled sequential frames $X^{\mathbb{T}}=\{(x^{\mathbb{T}}_{n,1},x^{\mathbb{T}}_{n,2}, \dots, x^{\mathbb{T}}_{n,k})\}_{n=1}^{N^{\mathbb{T}}}$ in the target domain, where $x^{\mathbb{T}}_{n,2}$ is an adjacent frame of $x^{\mathbb{T}}_{n,1}$, and $N^{\mathbb{T}}$ is the number of the target video sequences. 
Note that $x^{\mathbb{T}}\in \mathbb{R}^{H, W, 3}$, $x^{\mathbb{S}}\in \mathbb{R}^{\hat{H}, \hat{W}, 3}$, $y^{\mathbb{S}}\in \mathbb{B}^{\hat{H}, \hat{W}, C}$ as pixel-wise one-hot vectors, and $C$ is the number of all the categories $\mathbb{C}$ . 
For the geometric part, our motion network and the depth network are represented by $G_{m}$ and $G_{p}$. For the semantic part, our segmentation network is indicated by $G_{e}$. Our objective is to adapt the segmentation model $G_{e}$ to the unlabeled target domain.

\subsection{Target Domain Geometric Learning}
\label{geometric_learning}
The joint acquisition of knowledge concerning the moving objects and the motion of the ego camera within a local static scene has obtained considerable attention within the area of dynamic scene understanding~\cite{motion_net, 3dmotion_seokju, 3dmotion_2, cao_object_motion}. Recent investigations have introduced a method to disentangle the local objects' independent motion (called \textit{object motion}) from the global camera's motion (called \textit{ego-motion}) in a self-supervised manner with geometric constraints~\cite{motion_net,3dmotion_seokju}. We use the object motion to segment the moving objects out from the static scene.

Given an unlabeled target frame and its adjacent frame $\{x^{\mathbb{T}}_1, x^{\mathbb{T}}_2 \}$, our depth network $G_p$ is trained to estimate their depth maps $\{d^{\mathbb{T}}_1, d^{\mathbb{T}}_2\} \in  \mathbb{R}^{H,W}$. 
Then, these frames with their corresponding depth maps are concatenated as input $\{x^{\mathbb{T}}_1, d^{\mathbb{T}}_1, x^{\mathbb{T}}_2, d^{\mathbb{T}}_2\}$ and sent into the motion net $G_{m}$. Then $G_m$ is trained generates the camera's ego-motion $[R, t]$ (in 6 DoF) and the object motion $\Psi \in \mathbb{R}^{H,W,3}$ (in 3D space: $x$, $y$, and $z$-axis), where $R \in \mathbb{R}^{3}$ is the camera's ego rotation and $t\in \mathbb{R}^{3}$ is the camera's ego translation.  
On this basis, we use the camera's ego-motion $[R, t]$, the object motion $\Psi$, and the adjacent frame $x^{\mathbb{T}}_2$ to reconstruct the original image $x^{\mathbb{T}}_1$ with an inverse warping operation
\begin{equation}
    \hat{x}^{\mathbb{T}}_1 = \mathcal{F}(x^{\mathbb{T}}_2, d^{\mathbb{T}}_1, R, t, \Psi, K),
\end{equation}
where $ \hat{x}^{\mathbb{T}}_1$ is the reconstructed image by warping the adjacent image $x^{\mathbb{T}}_2$ (as reference), $\mathcal{F}$ is the projection operation using camera geometry~\cite{3dmotion_seokju}, and $K\in \mathbb{R}^{3\times 3}$ is the camera's intrinsic parameters. We adopt the photometric loss~\cite{3dmotion_2} and the regularization losses~\cite{motion_net,3dmotion_seokju} to optimize the motion network $G_m$ and the depth network $G_p$ together shown in Fig.~\ref{fig:geometric_training} (b).

\subsection{Motion Mask Preprocessing}
\label{motion mask preprocessing}
\begin{figure}[t!]
    \centering    \includegraphics[width=0.49\textwidth]{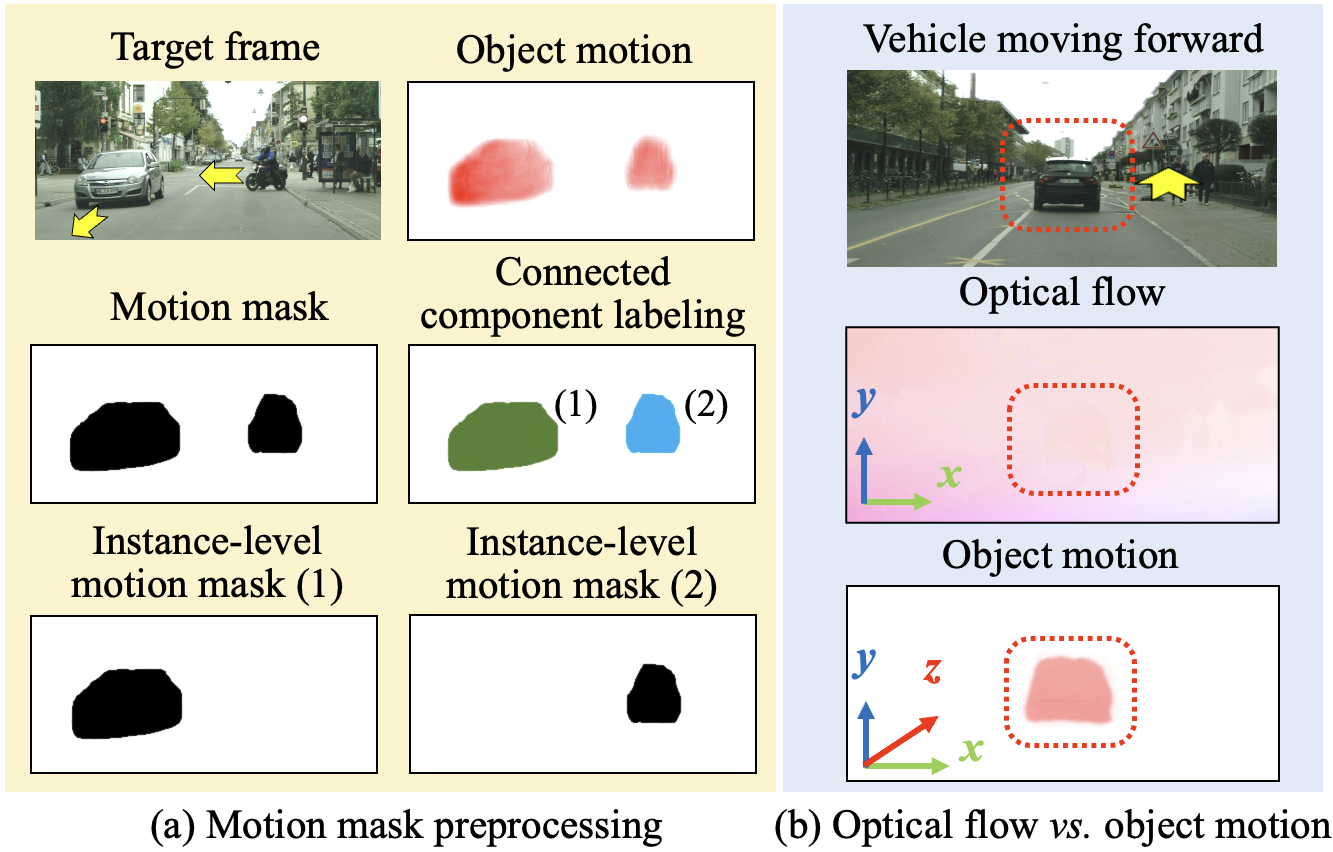}
    \caption{(a) The motion mask extracted from the object motion map include multiple moving instances. Therefore, we adopt connect component labeling to identify each moving instance. (b) The object motion map is in 3D space ($x, y, z$-axis). Object motion is capable to capture the motion pattern at $z$-axis (moving forward/backward) such as this vehicle. In contrast, optical flow which lies in 2D space fails to capture the motion of this vehicle.}
    \label{fig:motion_mask_preprocessing}
\end{figure}

Our motion mask preprocessing (MMP) aims at localizing the moving instances based on the object motion information in the target domain. An exemplary procedure of MMP is shown in Fig. \ref{fig:motion_mask_preprocessing} (a).
Given the motion network $G_m$ optimized by the photometric loss and the regularization losses, we first predict the object motion map $\Psi\in \mathbb{R}^{H,W,3}$ from the adjacent frames $x^{\mathbb{T}}_1, x^{\mathbb{T}}_2 \in \mathbb{R}^{H,W,3}$ as input. On this basis, we extract a binary motion mask $\Psi^{\mathbb{M}} \in \mathbb{B}^{H,W}$ from $\Psi$ via
\begin{equation}
\Psi^{\mathbb{M}}_{(i)}=
\left\{
\begin{array}{ll}
1, &  \text{if } |{\Psi}_{(i, d)}| >0,  \forall d \in \{1,2,3\} \\
0, &  \text{otherwise} 
\end{array}
 \right.,
\label{bmm}
\end{equation}
where $i$ indicate the pixel coordinate, $\Psi^{^{\mathbb{M}}}_{(i)}$ is the mask value on the image pixel $x^{\mathbb{T}}_{(i)}$. Note that the object motion $\Psi$ predicted $G_m$ is relative to the scene, \textit{e.g.}, it is separated from the camera's ego-motion. Therefore, we use $\Psi^{\mathbb{M}}$ to localize and segment all the moving objects in the scene. 

The binary motion mask $\Psi^{\mathbb{M}}$ could potentially include multiple moving instances, as it is common for multiple cars and motorcycles to move together within the same scenes.
To differentiate various moving instances, we utilize connected component labeling~\cite{connected_component} which is to identify each moving instance from $\Psi^{\mathbb{M}}$ with a unique label. 
We first run connected component labeling on $\Psi^{\mathbb{M}}$ to get a "new" map $\Tilde{\Psi}^{\mathbb{M}}$ with unique labels 
\begin{equation}
    \Tilde{\Psi}^{\mathbb{M}} = \Gamma (\Psi^{\mathbb{M}}),
\end{equation}
where $\Gamma$ is the connected component labeling process. For each unique label $m$ in $ \Tilde{\Psi}^{\mathbb{M}}$, we extract a binary motion mask $\psi^m \in \mathbb{B}^{H,W}$ for the $m\tp{th}$ moving instance. In this regard, we generate a set of binary instance-level motion masks $\{\psi^m\}_{m=1}^{M}$ via
\begin{equation}
    \{\psi^m\}_{m=1}^{M} = \Delta(\Tilde{\Psi}^{\mathbb{M}}),
    \label{eq:instance_level_motion_mask}
\end{equation}
where $\Delta$ denotes the instance-level motion mask extraction mentioned above (an example is shown in Fig.~\ref{fig:motion_mask_preprocessing} (a)), and $M$ is the number of the moving instance masks in $\Psi^{\mathbb{M}}$. 

\noindent \textbf{Difference with Optical Flow.} The object motion differs from the optical flow in two aspects. 1) Optical flow is not accurate for detecting the motion pattern on the front-to-back axis ($z$-axis) which is a common motion pattern in the real world, as it is a motion representation in \textit{2D space}. However, this issue doesn't exist in object motion which lies in \textit{3D space}. 2) Optical flow is a motion representation of all the pixels with respect to the camera's movement. Therefore, optical flow is a mixed motion representation of the object motion and ego-motion. In contrast, Object motion~\cite{motion_net,3dmotion_seokju} represents the independent movement of the objects, which is disentangled from the camera's ego-motion through the learning process in Sec.~\ref{geometric_learning}. Provided a car moving on the $z$-axis, we visualize the optical flow and the object motion in Fig.~\ref{fig:motion_mask_preprocessing} (b), where the object (the vehicle's) motion and camera's ego-motion are similar (toward the $z$-axis with similar velocity). The object motion successfully detects the the vehicle's motion at the $z$-axis. However, the optical flow cannot easily detect it because the the vehicle's motion is similar to the camera's ego-motion.

\section{Methodology}
MoDA consists of two modules: object discovery and semantic mining. The object discovery module takes the instance-level motion masks and extracts the moving object masks (in Sec.~\ref{subsec:object_discovery}). Subsequently, the semantic mining module utilizes the object masks to refine target pseudo labels (in Sec.~\ref{subsec:semantic_mining}).

\subsection{Segmentation Warm-up}
\label{subsec:segmentation_warm-up}
We first conduct warm-up training for segmentation network following DACS \cite{DACS}. Given a labeled source frame $x^{\mathbb{S}}$ and its ground-truth map $y^{\mathbb{S}}$, we first train the segmentation network $G_{e}$ with the supervised segmentation loss 
\begin{equation}
    \mathcal{L}_{ce}^{\mathbb{S}} = - \sum_{i=1}^{\hat{H},\hat{W}} \sum_{c=1}^{C} y_{(i,c)}^{\mathbb{S}} \log (p_{(i,c)}^{\mathbb{S}} ),
\end{equation}
where the source prediction $p^{\mathbb{S}} = G_{e}(x^{\mathbb{S}})\in \mathbb{R}^{\hat{H},\hat{W},C}$, $p_{(i,c)}^{\mathbb{S}}$ denotes the predicted softmax possibility on $c\tp{th}$ category on the pixel $x^{\mathbb{S}}_{(i)}$, 
$G_{e}$ is the segmentation network, and $C$ is the total number of categories.
The segmentation network trained solely on the source domain lacks generalization when applied to the target domain. To bridge the domain gap, we optimize the cross-entropy loss with the target pseudo labels. For simplicity, let the target pseudo label at the $t\tp{th}$ iteration denote as $\tilde{y}^{\mathbb{T}} \in \mathbb{B}^{H,W,C}$. The cross-entropy loss is defined by 
\begin{equation}
    \mathcal{L}_{ce}^{\mathbb{T}} = - \sum_{i=1}^{H,W} \sum_{c=1}^{C} \tilde{y}_{(i,c)}^{\mathbb{T}} \log (p_{(i,c)}^{\mathbb{T}} ).
    \label{eq:target self-training}
\end{equation}
During warm-up stage, we also follow the \textit{mixing augmentation between source and target samples} used in DACS.

\subsection{Target Object Discovery}
\label{subsec:object_discovery}
\begin{figure}[t!]
    \centering
    \includegraphics[width=.49\textwidth]{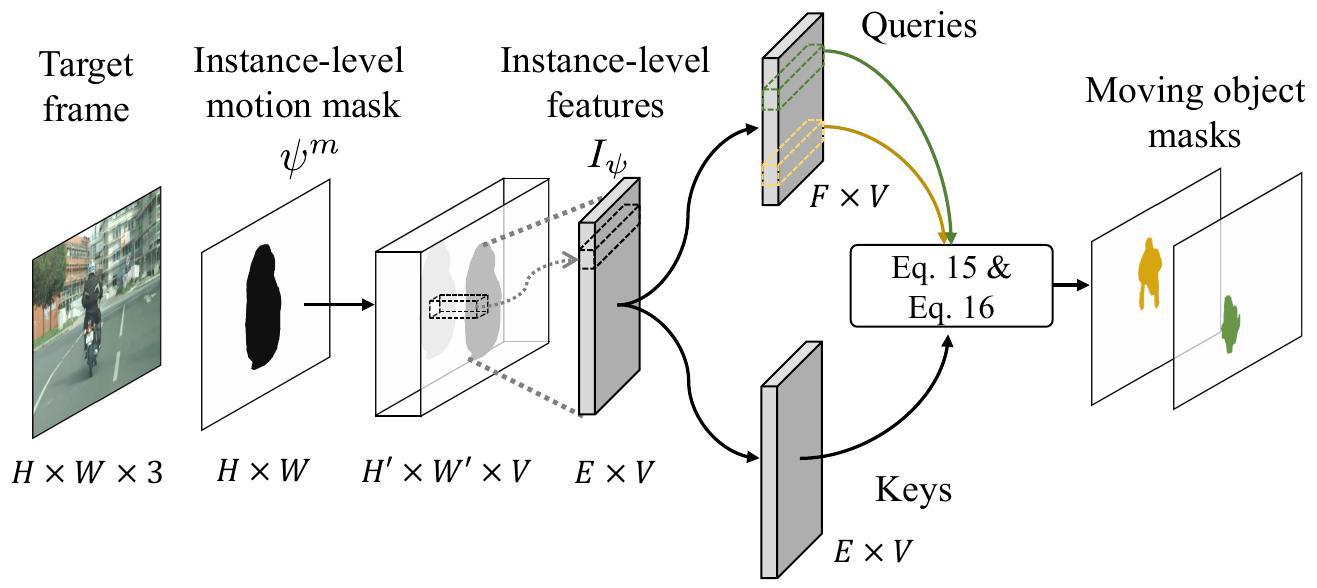}
    \caption{The instance-level motion mask might contain multiple moving objects bound together such as the \textit{rider} and the \textit{motorcycle}. The object discovery module takes an instance-level motion mask as input and predicts accurate object masks. Specifically, given a target image and its instance-level motion masks, we compute an objectness score map by computing the similarity of each query with all the keys in Eq.~\ref{eq:cosine_similarity} and the processing in Eq.~\ref{eq:norm_rank_nms}.}
    \label{fig:self_supervised_object_discovery}
\end{figure}
Directly applying the instance-level motion masks $\{ \psi^{m}\}_{m=1}^{M}$ for boosting the quality of target pseudo labels encounters two points of challenge. 1) The instance-level motion masks provide a \textit{coarse} segmentation of the moving objects due to the side effects of the motion regularization loss. Therefore, directly using the instance-level motion masks leads to noisy pseudo labels which might affect the performance of domain alignment. 2) There are some special cases where some moving instances might contain \textit{multiple} objects. For example, a motorcycle and its rider (two objects) are bounded into one moving instance (presented in Fig.~\ref{fig:self_supervised_object_discovery}).

To tackle these two issues, we propose a self-supervised object discovery module (ODM) to learn the accurate moving object masks from the instance-level motion masks $\{ \psi^{m}\}_{m=1}^{M}$, as shown in Fig.~\ref{fig:self_supervised_object_discovery}. First, given the target frame $x^{\mathbb{T}}$, a dense feature map $I\in \mathbb{R}^{H', W', V}$ is extracted from the segmentation network pre-trained on the source domain (shown in Sec.~\ref{subsec:segmentation_warm-up}). 
Then we adopt a self-supervised design to promote the objectness in the features' attention. Given an instance-level motion mask $\psi^{m} \in \mathbb{B}^{H,W}$ of $x^{\mathbb{T}}$ (generated by Eq.~\ref{eq:instance_level_motion_mask}), we bilinearly downsample $\psi$ to the spatial size of $I$ and select all the instance-level features that are covered by $\psi^{m}$, denoted by $I_{\psi} \in \mathbb{R}^{E,K}$, which is computed by
\begin{equation}
    I_{\psi} =\Upsilon \big(I \odot \texttt{repmat}(\texttt{bd}(\psi^{m})) \big),
\end{equation}
where $\texttt{bd}$ represents the binear downsampling, $\texttt{repmat}$  indicates the repeating operation that makes the shape of $\psi^{m}$ to be same as $I$, $\odot$ is the element-wise production, and $\Upsilon$ is to select all non-zero feature vectors.

To generate the binary moving object masks, we construct the queries and the keys from $I_{\psi}$. Our queries $Q\in \mathbb{R}^{F,V}$ are generated by a bilinear downsampling of $I_{\psi}$ where $F$ is the downsampled size, and our keys $K\in \mathbb{R}^{E, V}$ are from $I_{\psi}$ itself.  Given a query $Q_e \in \mathbb{R}^{V}$ in $Q$, we calculate its cosine similarity with all keys in $K$. Thus, we produce an \textit{objectness} score map $S\in \mathbb{R}^{E,F}$ by
\begin{equation}
    S_{e,f} = \texttt{cosim}(Q_e, K_f),
    \label{eq:cosine_similarity}
\end{equation}
where $K_f\in \mathbb{R}^{K}$ is the $f\tp{th}$ key of $K$, and $\texttt{cosim}$ represents the cosine similarity which is the dot product of two vectors with $\mathcal{L}_2$ normalization. Next, the \textit{objectness} score map is transformed by a normalization into a soft map where the scores are adjusted into the range of $[0, 1]$. To extract the binary moving object masks, a threshold value $\tau$ is applied to the soft map. The resulting binary moving object masks are ranked based on their objectness scores, and any redundant masks are eliminated through non-maximum-suppression (NMS). The entire procedure to get the moving object masks $ \{o^{j}\}_{j=1}^{J} $ is represented by 
\begin{equation}
    \{o^{j}\}_{j=1}^{J} = \texttt{NMS} \big( \texttt{rank} (\texttt{norm} (S)) \big),
    \label{eq:norm_rank_nms}
\end{equation}
where $o^{j} \in \mathbb{B}^{H,W}$ and $J$ is the number of the object masks predicted from $\psi^{m}$. 

\subsection{Target Semantic Mining}
\label{subsec:semantic_mining}

\begin{figure}[t!]
    \centering
    \includegraphics[width=0.5\textwidth]{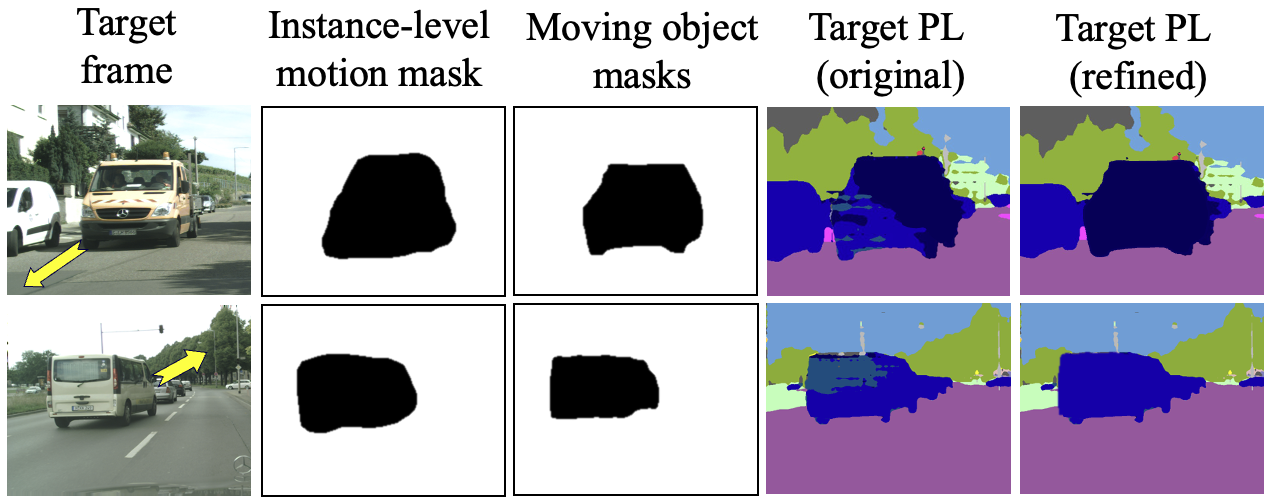}
    \caption{Directly utilizing instance-level motion masks might be sub-optimal as they are coarse masks for moving objects. The object discovery module is proposed to extract accurate object masks from coarse instance-level motion masks. The semantic mining module takes the moving object masks as guidance to refine the target pseudo labels.}
    \label{fig:target_semantic_mining}
\end{figure}

Our semantic mining module (SMM) takes the moving object masks as guidance to refine the noisy target pseudo labels. SMM is based on the assumption of \textit{rigidity of the moving objects}, \textit{e.g.}, vehicles, and motorbikes on traffic roads. For example, if a vehicle is moving, all parts of the vehicle are moving together. Based on the rigidity of the moving objects, all the pixels covered by a moving object mask in an image must have the same categorical label. Subsequently, we have the following remark: \\

\noindent \textbf{Remark 1.}\textit{ If a moving object mask is present, then the image pixels that it covers should have a semantic label that corresponds to the same moving categorical label.}  \\

Given a  target pseudo label $\tilde{y}^{'\mathbb{T}}$, we choose a dominant category $c^{*}$ in $\tilde{y}^{'\mathbb{T}}$ that are covered by the moving object mask $o^{j}$. Concretely, $c^{*}$ is determined by the moving category with the highest occurrence. Based on \textbf{Remark 1}, we introduce a semantic mining weight $w\in\mathbb{R}^{H,W,C}$ to update the target pseudo label via
\begin{equation}
w_{(i,c)} = \lambda o^{j}_{(i)} \mathbbm{1}(c=c^{*}),
\label{eq:semantic mining weight}
\end{equation}
where $\mathbbm{1}(\cdot)$ is the indicator function, $o^{j}_{(i)}$ is the object mask value on the pixel $x_{(i)}^{\mathbb{T}}$, and $\lambda$ is a non-negative hyperparameter for weighting. Then, we update the target pseudo label using the following equation
\begin{equation}
    \tilde{y}^{'\mathbb{T}} = \Gamma(\texttt{softmax}(( w+1)\odot p^{\mathbb{T}} )),
\end{equation}
where $\tilde{y}^{'\mathbb{T}}$ is the updated target pseudo label by our target semantic mining, and $\Gamma$ is the process of taking the most probable category from $p^{\mathbb{T}}$. We provide an illustration of using object motion masks to update noisy target pseudo labels shown in Fig.~\ref{fig:target_semantic_mining}. We use the updated target pseudo label to update the segmentation network $G_{e}$ by $\mathcal{L}_{SMM}$ which is defined by
\begin{equation}
     \mathcal{L}_{MoDA} = \sum_{i=1}^{H,W} \sum_{c=1}^{C} \tilde{y}_{(i,c)}^{'\mathbb{T}} \log \big( G_{e}(x^{\mathbb{T}}) \big).
     \label{eq:semantic_mining_module}
\end{equation}
\begin{figure}[t!]
    \centering
    \includegraphics[width=0.5\textwidth]{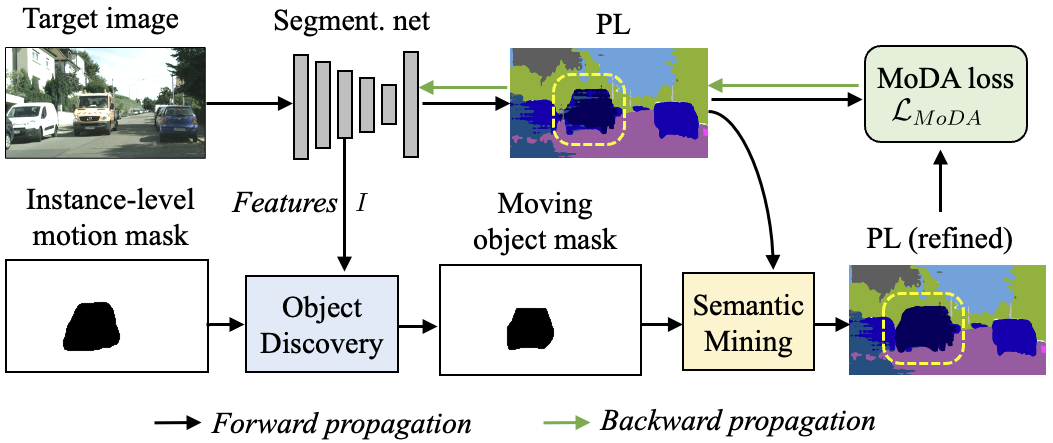}
    \caption{MoDA utilizes object motion information (instance-level motion masks) as cues to refine target pseudo labels. The key components of MoDA are the object discovery and the semantic mining module, which do not require any training. The object discovery module takes instance-level motion masks and extracts the moving object masks. These moving object masks are then sent as input to the semantic mining module to refine the target pseudo labels (PL). Note that the pseudo labels (PL) are generated by the segmentation network pre-trained at the warm-up stage.}
    \label{fig:overall_architecture}
\end{figure}
The overview of MoDA is shown in Fig.~\ref{fig:overall_architecture}. \\


\noindent \textbf{Optical Flow Regularization (OFR).} Our baseline is to use the optical flow information from the adjacent target frames. Specifically, we can propagate the prediction of the previous frame to the current frame using optical flow estimates between the frames and subsequently ensure the consistency between the prediction of the current frame and the propagated prediction from the previous frame. Given two adjacent frames $\{x^{\mathbb{T}}_1, x^{\mathbb{T}}_2 \}$, we forward them as input to get the predictions $\{ p^{\mathbb{T}}_1, p^{\mathbb{T}}_2 \}$. Moreover, we use FlowNet~\cite{ilg2017flownet} to estimate the optical flow $f^{\mathbb{T}}_{1\to 2}$ from $x^{\mathbb{T}}_1$ to $x^{\mathbb{T}}_2$. Then we warp the prediction $ p^{\mathbb{T}}_1$ to the propagated prediction $ \bar{p}^{\mathbb{T}}_2$. Then the optical flow regularization loss $\mathcal{L}_{OFR}$ is formulated as
\begin{equation}
    \mathcal{L}_{OFR} = \|  p^{\mathbb{T}}_1 -  \bar{p}^{\mathbb{T}}_2 \|_2.
    \label{ofr}
\end{equation}
We conduct experiments and compare MoDA with the baseline method using optical flow regularization $\mathcal{L}_{OFR}$ that considers the temporal consistency of the target frames in Sec.~\ref{sec:experiments}.


\section{Experiments}
\label{sec:experiments}
The datasets and implementation details are in Sec.~\ref{experiment setup}. The evaluation is conducted on domain adaptive video segmentation and image segmentation task in Sec.~\ref{subsec:evaluation_results}. The ablation study and hyperparameter analysis are presented in Sec.~\ref{ablation study}.

\subsection{Experiment Setup}
\label{experiment setup}
\noindent \textbf{Datasets.} For \textit{domain adaptive image segmentation}, we have GTA5~\cite{GTA5} as the source domains. GTA5 contains 24,966 training images with a resolution $2,048\times1,024$ and $19$ categories. For \textit{domain adaptive video segmentation}, we adopt VIPER~\cite{VIPER} as the source domain. VIPER~\cite{VIPER} contains $133,670$ synthetic frames with the corresponding pixel-wise annotations from $77$ videos generated from game engines. For the target domain, we use Cityscapes-Seq~\cite{DA-VSN,TPS} which contains 500 video sequences from the Cityscapes dataset~\cite{cityscapes}, and each sequence consists of 30 frames.  We also include 500 validation images from the Cityscapes dataset for evaluation. \\

\noindent \textbf{Implementation details.} We conduct experiments on two types of architectures: CNN-based architecture and Transformer-based architecture. For CNN-based architectures, our warm-up stage follows DACS~\cite{DACS}. We first adopt DeepLab-V2~\cite{deeplabv2} with ReseNet-101~\cite{resnet} for the segmentation network, pre-trained on ImageNet~\cite{imagenet}. For Transformer-based architecture, we adopt MiT-B5~\cite{segformer} as the encoder and incorporate our MoDA with existing state-of-the-art approaches~\cite{daformer,HRDA} by using the pre-trained weights from them. Our batch size is 16 with 8 source and 8 target images with the resolution $1,024\times512$. Threshold $\tau$ is set with $0.5$. The optimizer for segmentation is SGD~\cite{SGD} with learning rate of $2.5e^{-4}$, momentum $0.9$, and weight decay of $5\times 10^{-4}$. We optimize the discriminator using Adam with the initial learning rate of $10^{-4}$. For the momentum network, we set $\lambda=0.99$. We implement MoDA with PyTorch and the training process is running on two Titan RTX A6000 GPUs. 

\begin{table*}[t!]
\caption{The comparison with baseline methods on {domain adaptive \textit{video} segmentation} benchmark VIPER$\to$Cityscapes-Seq and {domain adaptive \textit{image} segmentation} benchmark GTA5$\to$Cityscapes-Seq. Current baseline methods are optimized with optical flow regularization (represented by \textit{+OFR}) on the target frames. MoDA utilizes object motion as cues to guide domain adaptation and demonstrates superior performance against current baseline methods using optical flow regularization. This is to demonstrate that \textit{object motion serves as stronger guidance for domain adaptation compared to optical flow}. We also show MoDA is \textit{versatile} as it complements existing state-of-the-art domain adaptive image segmentation approach (\textit{+MoDA}).}
\vspace{-5pt}
\centering
\resizebox{0.9\textwidth}{!}{
\begin{tabular}{r | c c c c c c c c c c c c c c c | c} 
\multicolumn{17}{c}{VIPER $\to$ Cityscapes-Seq} \\ 
\hline
Method & Road & Side. & Buil. & Fence & TL & TS & Vege. & Terr. & Sky & Pers. & Car & Truck & Bus & Motor & Bike & mIoU \\ \hline

\multicolumn{17}{l}{\textit{Backbone: ResNet-101}} \\ \hline

AdvEnt + OFR~\cite{advent} &78.6&30.0&79.9&23.9&27.3&28.1&82.2&13.0&81.2&59.5&62.3&6.4&40.3&4.8&2.7&41.3 \\

CBST + OFR ~\cite{CBST}&48.0&20.9&\textbf{85.6}&12.4&19.2&21.0&82.6&19.5&{83.2}&60.0&71.8&3.9&39.5&23.1&{38.5}&41.9 \\

IntraDA + OFR~\cite{IntraDa}&80.4&35.1&80.7&\textbf{24.7}&28.2&{24.7}&82.5&14.0&79.8&60.0&63.1&6.0&41.8&6.4&4.2&42.1 \\

CRST + OFR~\cite{CRST}&55.5&22.0&82.4&12.8&19.9&16.0&85.8&19.3&83.1&62.6&72.4&5.0&39.8&\textbf{29.6}&34.9&42.7 \\

DACS + OFR~\cite{DACS} &71.2&26.8&83.5&20.7&23.3&24.1&82.9&{24.9}&81.8&58.4&76.3&28.8&{41.3}&{24.7}&22.8&46.1\\

PixMatch + OFR~\cite{PixelMatch} & 78.8 & 28.4 & 82.2 & 18.2 & \textbf{30.8} & 25.3 & 84.6 & \textbf{31.4} & \textbf{83.3}
& 59.1 & 75.2 & 34.3 & 43.7 & 15.9 & 13.1 & 46.9 \\

DA-VSN~\cite{DA-VSN} & \textbf{87.1} & \textbf{38.3}& 82.2 & {23.7} & {29.8} & 28.4 & \textbf{85.9} & 26.6 & 80.8 & 60.3 & 78.7 & 21.7 & 46.9 & 22.0 & 10.5 & 48.2 \\  

TPS~\cite{TPS} & 83.4 & 35.8 & 78.9 & 9.6 & 25.7 & \textbf{29.5} & 77.9 & {28.4} & 81.6 & 60.2 & 81.1 & 40.7 & 39.8 & 27.7 & 31.4 & 48.7 \\ 

\rowcolor{BA}
\textbf{MoDA} & {72.2} &{25.9}&{80.9} & 18.3 & 24.6 & {21.1} & 79.1 & 23.2 & {78.3} &\textbf{68.7} & \textbf{84.1 }& \textbf{43.2} &\textbf{49.5}& {28.8}& \textbf{38.6}& \textbf{49.1}  \\ 

\hline
\end{tabular}
}
\centering
\resizebox{\textwidth}{!}{
\begin{tabular}{r | c c c c c c c c c c c c c c c c c c c | c }
\multicolumn{21}{c}{GTA5 $\to$ Cityscapes-Seq} \\ 
\hline
Method & Road & Side. & Buil. & Wall & Fence & Pole & TL & TS & Vege. & Terr. & Sky & Pers. & Rider & Car & Truck & Bus & Train & Motor & Bike & mIoU \\ \hline
\multicolumn{21}{l}{\textit{Backbone: ResNet-101}} \\ \hline
AdatpSegNet + OFR~\cite{adaptsegnet}&86.3&36.2&79.8&24.1&23.9&24.1&36.0&15.2&82.6&31.9&74.4&58.7&26.8&75.1&33.9&35.8&4.1&29.7&28.8&42.5 \\
AdvEnt + OFR~\cite{advent}&91.6&54.0&79.8&32.4&21.5&33.6&29.1&21.7&84.2&35.4&81.7&52.9&23.8&81.9&31.0&35.3&16.8&26.2&\textbf{43.7}&46.1  \\
IntraDA + OFR~\cite{IntraDa}&91.6&38.1&81.7&33.0&20.4&28.6&31.9&22.7&85.6&41.1&78.9&59.2&31.8&86.1&31.9&48.3&0.2&30.9&35.7&46.2 \\
CRST+ OFR~\cite{CRST}&90.0&56.1&83.3&32.8&24.5&36.6&33.7&25.4&84.9&35.3&81.0&58.3&25.6&84.0&28.7&31.8&27.4&25.8&42.9&47.8 \\
SIM+ OFR~\cite{SIM}&89.8&46.1&{85.9}&33.5&27.8&36.7&35.0&37.6&84.7&\textbf{45.4}&83.2&56.4&31.5&82.7&43.2&49.9&2.1&33.4&38.7&49.7  \\
CAG-UDA + OFR~\cite{CAG_UDA} &91.2&50.3&84.1&37.5&26.9&37.6&26.2&49.3&86.2&37.9&77.4&55.8&37.9&86.2&20.1&43.2&\textbf{44.1}&28.6&36.5&50.4 \\
IAST + OFR~\cite{IAST}&\textbf{92.5}&\textbf{59.3}&84.2&\textbf{40.7}&25.8&27.3&42.8&36.5&82.7&31.6&89.5&61.7&30.3&87.6&41.4&46.7&28.9&33.0&42.4&51.8 \\
DACS + OFR~\cite{DACS}&90.4&40.1&\textbf{86.6}&31.4&\textbf{40.8}&39.2&\textbf{45.8}&53.4&88.6&42.7&89.5&66.1&35.7&85.6&46.7&51.3&0.2&28.7&34.6&52.5 \\
\rowcolor{BA}
\textbf{MoDA}&91.7&43.2&85.3&31.7&39.6&\textbf{40.5}&44.3&\textbf{54.8}&\textbf{89.7}&41.8&\textbf{90.4}&\textbf{68.5}&\textbf{41.4}&\textbf{87.9}&\textbf{48.1}&\textbf{56.7}&11.8&\textbf{35.3}&39.6&\textbf{54.9} \\
\hline

\multicolumn{21}{l}{\textit{Backbone: Transformer}} \\ \hline

HRDA~\cite{HRDA}&{97.1}& \textbf{74.5} &90.8&\textbf{62.2}&{51.0}& 58.2 &63.4&\textbf{70.5}&\textbf{92.0}&48.3&94.6& 78.8 & 53.1 & 94.5& \textbf{83.6}& 84.9 & {76.8} &64.3&65.7&73.9 \\
\rowcolor{BA}
HRDA\textbf{ + MoDA}&\textbf{97.3} &{74.4} & \textbf{91.4} &{61.4} & \textbf{51.4} & \textbf{59.2} & \textbf{64.1} & 70.0 & 91.0 &\textbf{49.5} &\textbf{95.9} &\textbf{80.1} & \textbf{57.1} & \textbf{95.1} & 83.1 &\textbf{89.4} &\textbf{77.5} & \textbf{72.0} &\textbf{68.8} &\textbf{75.2} \\ \hline 

\end{tabular}
}

\label{tab: GTA52City_Seq}
\end{table*}

\begin{table}[t!]
    \centering
    \caption{Ablation study on the components of MoDA: object discovery module (ODM) and semantic mining module (SMM).}
    \resizebox{0.45\textwidth}{!}{
    \begin{tabular}{r | c c c | c c}
    \hline
    Configuration        & Warm-up    & SMM       &      ODM    &     mIoU    &     Gap \\ \hline
    Only Warm-up         & \checkmark &           &             &        45.9 &     -3.2 \\
     w/o Semantic Mining & \checkmark &           & \checkmark  &        45.9 &    -3.2 \\
    w/o Object Discovery & \checkmark & \checkmark &            &     46.7    &    -2.4 \\
    \rowcolor{BA} Full Framework (\textbf{MoDA}) & \checkmark & \checkmark    & \checkmark  & 49.1   & - \\
    \hline
    \end{tabular}
    }
    
    \label{tab:components of moda}
\end{table}

\subsection{Evaluation Results}
\label{subsec:evaluation_results}
\begin{figure}[t!]
    \centering
    \includegraphics[width=0.46\textwidth]{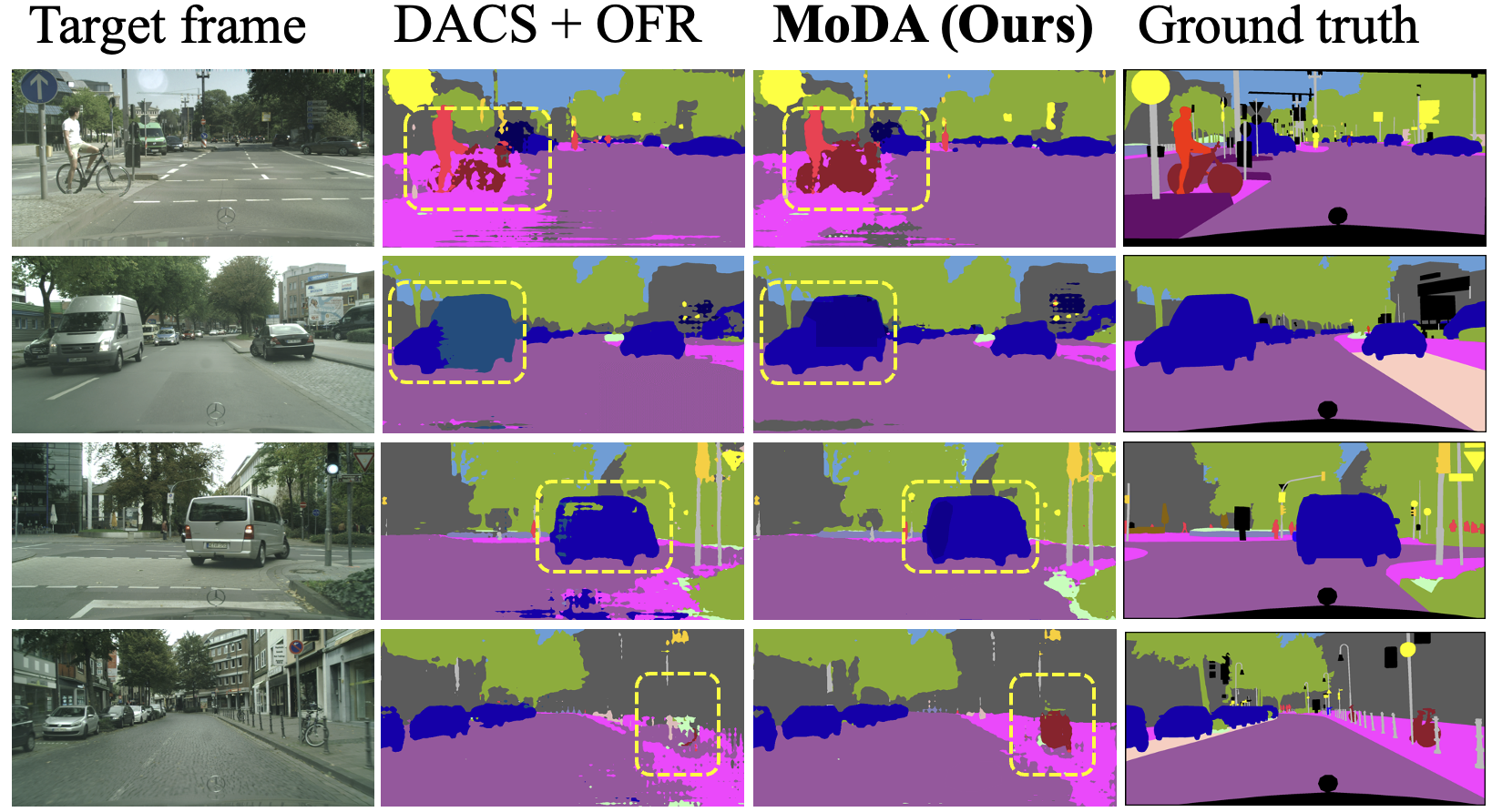}
    \caption{ Comparison of the qualitative results generated from MoDA and the baseline model on VIPER$\to$Cityscapes-Seq benchmark. }
    \label{fig:more_results}
\end{figure}

\begin{figure}
    \centering
    \includegraphics[width=0.46\textwidth]{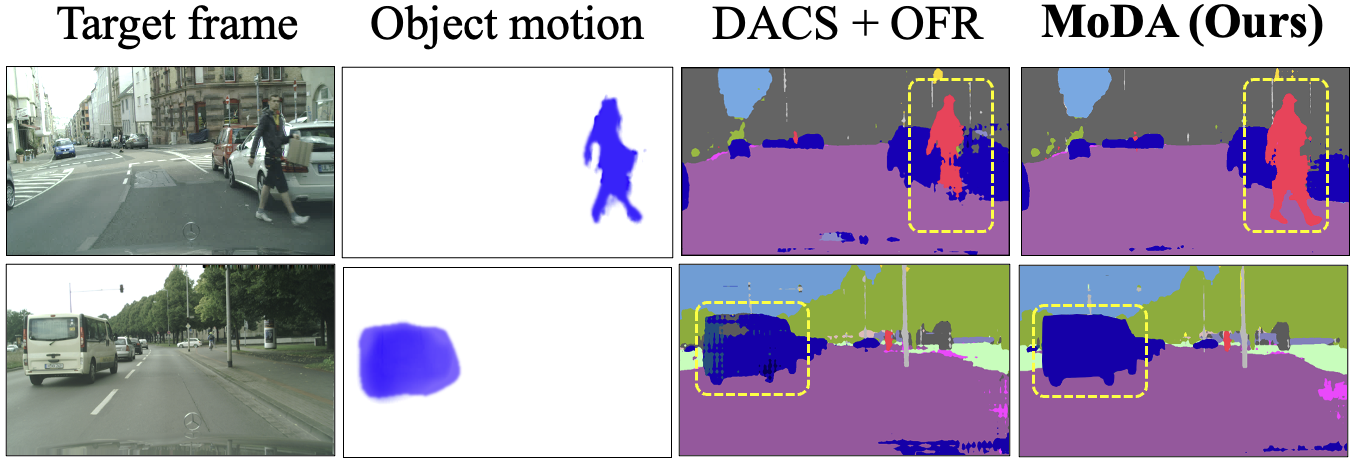}
    \caption{MoDA utilizes the object motion as cues to guide domain adaptation for segmentation task. MoDA outperforms the baseline model DACS + OFR~\cite{DACS} which adopts the optical flow regularization. This is to show that the object motion information is a strong cue to guide adaptation compared with optical flow information.}
    \label{fig:more_results_1}
\end{figure}

\noindent \textbf{Domain Adaptive Video Segmentation.}
We evaluate MoDA in the setting of domain adaptive video segmentation: \textit{VIPER$\to$Cityscapes-Seq}. Our baseline approaches are DA-VSN~\cite{DA-VSN} and TPS~\cite{TPS} which have included optical flow regularization. We also include domain adaptive image segmentation baselines: AdvEnt~\cite{advent}, CBST~\cite{CBST}, IntraDA~\cite{IntraDa}, CRST~\cite{CRST}, PixMatch~\cite{PixelMatch}, and DACS~\cite{DACS}. To make a fair comparison, all these domain adaptive image segmentation baselines are optimized with optical flow regularization ($\mathcal{L}_{OFR}$ in Eq.~\ref{ofr}). For example, DACS + OFR represents the training results of DACS~\cite{DACS}  with optical flow regularization. The experimental results are shown in Table~\ref{tab: GTA52City_Seq}. MoDA outperforms the baseline DACS+OFR, which shows that \textit{object motion is stronger guidance for domain adaptation compared with optical flow information}. Specifically, MoDA outperforms DA-VSN and TPS, indicating that \textit{self-supervised object motion from unlabeled video sequences are important information to be considered in video domain adaptation setting}. We provide qualitative examples of MoDA and the baseline DACS+OFR in Fig.~\ref{fig:more_results}. We also visualize examples of object motion maps learned from the motion network as cues to refine target pseudo labels in Fig. \ref{fig:more_results_1}.  \\ 

\noindent \textbf{Domain Adaptive Image Segmentation.} We include existing ResNet-101 based approaches for comparison: AdaptSegNet~\cite{adaptsegnet}, AdvEnt~\cite{advent}, IntraDA~\cite{IntraDa}, SIM~\cite{SIM}, CRST~\cite{CRST}, CAG-UDA~\cite{CAG_UDA}, IAST~\cite{IAST}, DACS~\cite{DACS}, and HRDA~\cite{HRDA}. We evaluate the performance of MoDA in Table~\ref{tab: GTA52City_Seq} ({GTA5 $\to$ Cityscapes-Seq}). To make a fair comparison, all the domain adaptive image segmentation baselines are optimized with {optical flow regularization} ($\mathcal{L}_{OFR}$ in Eq.~\ref{ofr}). Experimental results on Table~\ref{tab: GTA52City_Seq} demonstrate that MoDA outperforms existing domain adaptive image segmentation methods with optical flow regularization. This is to show that \textit{the object motion information is a strong guidance for domain adaptation compared with optical flow information}. We also combine MoDA with current state-of-the-art approaches and the results are showing with \textbf{+MoDA}. For example, the results of HRDA\textbf{+MoDA} are obtained by using HRDA~\cite{HRDA} as the warm-up stage to generate target pseudo labels and refine these pseudo labels with MoDA. This is to show that MoDA \textit{complements existing state-of-the-art UDA approach}. \\
\vspace{-10pt}


\subsection{Ablation Study}
\label{ablation study}

\begin{table}[t!]   
    \caption{We compare object motion with optical flow regularization after the same warm-up on VIPER$\to$Cityscapes-Seq benchmark. MoDA adopts DACS as the warm-up step and utilizes object motion as cues to refine target pseudo labels. DACS+OFR uses DACS as the warm-up step and applies optical flow regularization. This is to show that object motion functions as a stronger guidance compared with optical flow.}
    \centering
    \resizebox{0.45\textwidth}{!}{
    \begin{tabular}{r| c c c c |c  c}   
        \hline
        Configuration & $\mathcal{L}_{ce}^{\mathbb{S}}$  & $\mathcal{L}_{ce}^{\mathbb{T}} $ & $\mathcal{L}_{MoDA}$  & $\mathcal{L}_{OFR}$ & mIoU & \textit{Gain} \\
        \hline
        DACS~\cite{DACS} & \checkmark & \checkmark &  & & 45.9 & - \\
        DACS + OFR & \checkmark & \checkmark &  & \checkmark & 46.1 & +0.2\\
        \rowcolor{BA} \textbf{MoDA} &\checkmark & \checkmark& \checkmark & & 49.1 & +3.2\\
        \hline
    \end{tabular}
    }
    \label{tab: temporal consistency regularization}
\end{table}
\noindent \textbf{Optical Flow Regularization.} We provide an ablation study using object motion as guidance in comparison with the optical flow regularization. We show the evaluation of VIPER$\to$Cityscapes-Seq in Table~\ref{tab: temporal consistency regularization}. DACS+OFR achieves $46.1\%$ of mIoU by using optical flow regularization (OFR) which is $0.2\%$ higher than the DACS model~\cite{DACS}. MoDA produces $49.1\%$ of mIoU which is higher than DACS+OFR. This is to show that object motion functions as a stronger guidance compared with optical flow. \\

\noindent \textbf{Different Components in MoDA.} We conduct an ablation study on the effectiveness of the object discovery module (ODM) and semantic mining module (SMM) in Table~\ref{tab:components of moda}. On VIPER$\to$Cityscapes-Seq benchmark, by only using the warm-up step (without using ODM or SMM), we get the score $45.9\%$ of mIoU. On the other hand, utilizing the warm-up along with SMM (without using ODM), MoDA experienced a significant performance drop to $46.7\%$ of mIoU. It is caused by two reasons: \textit{1) The instance-level motion masks are not accurate to be used as the object masks in the semantic mining module; 2) Some instance-level motion masks contain multiple moving objects bounded together such as a motorcycle with a rider}.\\


\begin{figure}[t]
    \centering
    \includegraphics[width=0.47\textwidth]{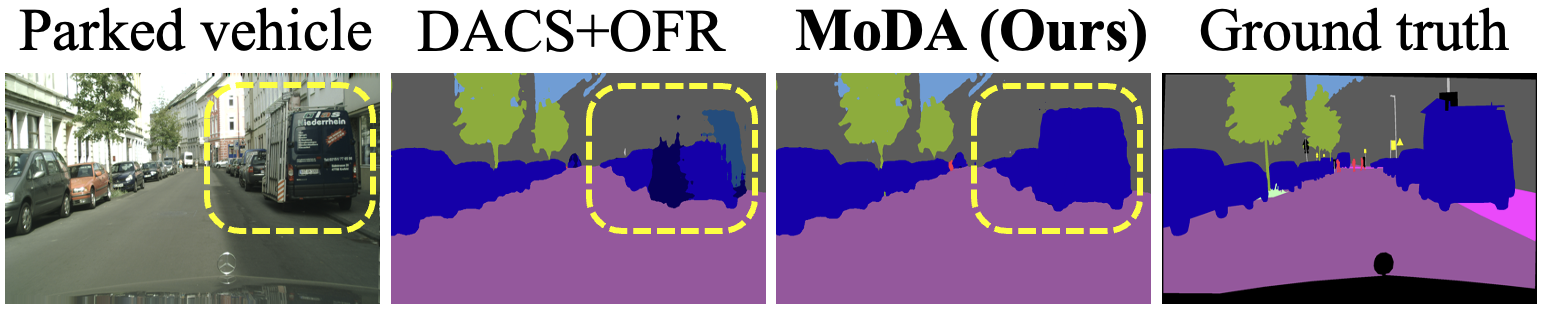}
    \caption{MoDA utilizes object motion information as guidance for domain adaptation. It is effective both for the moving and the static objects, such as the \textit{parked} vehicle in the yellow box. }
    \label{fig:parked_objects}
\end{figure}
\noindent \textbf{{What happens to the potentially movable, but static objects (\eg, parked cars, standing persons)?}} First of all, the overall training pipeline of our approach does not harm static objects like parked cars or standing pedestrians during the domain transfer. Since MoDA uses motion-guided object masks to update the noisy predictions of the pseudo labels, the performance on static objects will also be upgraded by updating the segmentation net with these new pseudo labels. As an example in Fig.~\ref{fig:parked_objects}, MoDA generates more accurate predictions on the parked vehicle that is not moving in comparison with the baseline DACS+OFR~\cite{DACS}. \\

\begin{table}[t!]   
    \caption{The ablation study on the hyperparameter $\lambda$ for weighting in semantic mining module.}
    \label{tab:lambda}
    \centering
    \resizebox{0.45\textwidth}{!}{
    \begin{tabular}{c|c c c c c c c c c }
    \hline
    $\lambda$ & 0.0 & 0.2 & 0.4 & 0.6 & 0.8 & 1.0 & 1.2 & 1.4 \\
    \hline
    mIoU & 45.9 & 52.9  & 53.4 & 54.1 & 49.1 & 49.1  & 49.1 & 49.1 \\
    \hline
    \end{tabular}
    }
\end{table}

\noindent \textbf{Hyperparameter $\lambda$.} We conduct an ablation study on the hyperparameter $\lambda$ in semantic mining module(Eq.~\ref{eq:semantic mining weight}). We present different values of $\lambda$ for the final performance in GTA5$\to$Cityscapes-Seq in Table~\ref{tab:lambda}. The bigger value of $\lambda$ puts more weight on semantic mining to update target pseudo labels. Our ablation results indicate that MoDA reaches the best performance when $\lambda$ reaches $0.8$. Additionally, it shows that MoDA's performance is insensitive to the hyperparameter $\lambda$ value when it is greater than $0.8$. \\

\vspace{-15pt}
\section{Conclusion}
This paper proposed a novel motion-guide domain adaption method, namely MoDA for the semantic segmentation task. MoDA utilizes the self-supervised object motion as cues to guide cross-domain alignment for segmentation task. MoDA consists of two modules namely, object discovery and semantic mining, to refine target pseudo labels. These refined pseudo labels are used in the self-training loop to bridge the cross-domain gap. On domain adaptive image and video segmentation experiments MoDA shows the effectiveness utilizing object motion as guidance for domain alignment compared with optical flow information. Moreover, MoDA is versatile as it complements existing state-of-the-art UDA approaches.
{
    \small
    \bibliographystyle{ieeenat_fullname}
    \bibliography{main}

\begin{thebibliography}{48}
\providecommand{\natexlab}[1]{#1}
\providecommand{\url}[1]{\texttt{#1}}
\expandafter\ifx\csname urlstyle\endcsname\relax
  \providecommand{\doi}[1]{doi: #1}\else
  \providecommand{\doi}{doi: \begingroup \urlstyle{rm}\Url}\fi

\bibitem[Araslanov and Roth(2021)]{da_sac}
Nikita Araslanov and Stefan Roth.
\newblock Self-supervised augmentation consistency for adapting semantic segmentation.
\newblock In \emph{IEEE Conf. Comput. Vis. Pattern Recog.}, pages 15384--15394, Virtual/Online, 2021.

\bibitem[Bottou(2010)]{SGD}
L{\'e}on Bottou.
\newblock Large-scale machine learning with stochastic gradient descent.
\newblock In \emph{Int. Conf. on Comput. Statis.}, pages 177--186, Paris, France, 2010. Springer.

\bibitem[Cao et~al.(2019)Cao, Kar, Hane, and Malik]{cao_object_motion}
Zhe Cao, Abhishek Kar, Christian Hane, and Jitendra Malik.
\newblock Learning independent object motion from unlabelled stereoscopic videos.
\newblock In \emph{IEEE Conf. Comput. Vis. Pattern Recog.}, 2019.

\bibitem[Chen et~al.(2017)Chen, Papandreou, Kokkinos, Murphy, and Yuille]{deeplabv2}
Liang-Chieh Chen, George Papandreou, Iasonas Kokkinos, Kevin Murphy, and Alan~L Yuille.
\newblock Deeplab: Semantic image segmentation with deep convolutional nets, atrous convolution, and fully connected crfs.
\newblock \emph{IEEE Trans. Pattern Anal. Mach. Intell.}, 40\penalty0 (4):\penalty0 834--848, 2017.

\bibitem[Cordts et~al.(2016)Cordts, Omran, Ramos, Rehfeld, Enzweiler, Benenson, Franke, Roth, and Schiele]{cityscapes}
Marius Cordts, Mohamed Omran, Sebastian Ramos, Timo Rehfeld, Markus Enzweiler, Rodrigo Benenson, Uwe Franke, Stefan Roth, and Bernt Schiele.
\newblock The cityscapes dataset for semantic urban scene understanding.
\newblock In \emph{IEEE Conf. Comput. Vis. Pattern Recog.}, pages 3213--3223, Las Vegas, Nevada, 2016.

\bibitem[Deng et~al.(2009)Deng, Dong, Socher, Li, Li, and Fei-Fei]{imagenet}
Jia Deng, Wei Dong, Richard Socher, Li-Jia Li, Kai Li, and Li Fei-Fei.
\newblock Imagenet: A large-scale hierarchical image database.
\newblock In \emph{IEEE Conf. Comput. Vis. Pattern Recog.}, pages 248--255, Miami, USA, 2009. Ieee.

\bibitem[Gao et~al.(2023)Gao, Cai, Bi, Li, Li, and Zheng]{electronics3}
Yuefang Gao, Yiteng Cai, Xuanming Bi, Bizheng Li, Shunpeng Li, and Weiping Zheng.
\newblock Cross-domain facial expression recognition through reliable global--local representation learning and dynamic label weighting.
\newblock \emph{Electronics}, 12\penalty0 (21):\penalty0 4553, 2023.

\bibitem[Gordon et~al.(2019)Gordon, Li, Jonschkowski, and Angelova]{3dmotion_2}
Ariel Gordon, Hanhan Li, Rico Jonschkowski, and Anelia Angelova.
\newblock Depth from videos in the wild: Unsupervised monocular depth learning from unknown cameras.
\newblock In \emph{Int. Conf. Comput. Vis.}, pages 8977--8986, Seoul, South Korea, 2019.

\bibitem[Guan et~al.(2021)Guan, Huang, Xiao, and Lu]{DA-VSN}
Dayan Guan, Jiaxing Huang, Aoran Xiao, and Shijian Lu.
\newblock Domain adaptive video segmentation via temporal consistency regularization.
\newblock In \emph{Int. Conf. Comput. Vis.}, pages 8053--8064, Virtual/Online, 2021.

\bibitem[Guizilini et~al.(2021)Guizilini, Li, Ambruș, and Gaidon]{GUDA}
Vitor Guizilini, Jie Li, Rareș Ambruș, and Adrien Gaidon.
\newblock Geometric unsupervised domain adaptation for semantic segmentation.
\newblock In \emph{Int. Conf. Comput. Vis.}, pages 8537--8547, Virtual/Online, 2021.

\bibitem[He et~al.(2016)He, Zhang, Ren, and Sun]{resnet}
Kaiming He, Xiangyu Zhang, Shaoqing Ren, and Jian Sun.
\newblock Deep residual learning for image recognition.
\newblock In \emph{IEEE Conf. Comput. Vis. Pattern Recog.}, pages 770--778, Las Vegas, Nevada, 2016.

\bibitem[Hoyer et~al.(2022{\natexlab{a}})Hoyer, Dai, and Van~Gool]{HRDA}
Lukas Hoyer, Dengxin Dai, and Luc Van~Gool.
\newblock Hrda: Context-aware high-resolution domain-adaptive semantic segmentation.
\newblock In \emph{Eur. Conf. Comput. Vis.}, pages 372--391, Tel Aviv, Israel, 2022{\natexlab{a}}. Springer.

\bibitem[Hoyer et~al.(2022{\natexlab{b}})Hoyer, Dai, and Van~Gool]{daformer}
Lukas Hoyer, Dengxin Dai, and Luc Van~Gool.
\newblock Daformer: Improving network architectures and training strategies for domain-adaptive semantic segmentation.
\newblock In \emph{IEEE Conf. Comput. Vis. Pattern Recog.}, pages 9924--9935, New Orleans, US, 2022{\natexlab{b}}.

\bibitem[Ilg et~al.(2017)Ilg, Mayer, Saikia, Keuper, Dosovitskiy, and Brox]{ilg2017flownet}
Eddy Ilg, Nikolaus Mayer, Tonmoy Saikia, Margret Keuper, Alexey Dosovitskiy, and Thomas Brox.
\newblock Flownet 2.0: Evolution of optical flow estimation with deep networks.
\newblock In \emph{IEEE Conf. Comput. Vis. Pattern Recog.}, pages 2462--2470, Hawaii, US, 2017.

\bibitem[Jia et~al.(2024)Jia, Tian, Hu, Jing, Zuo, and Li]{electronics4}
Longfei Jia, Xianlong Tian, Yuguo Hu, Mengmeng Jing, Lin Zuo, and Wen Li.
\newblock Style-guided adversarial teacher for cross-domain object detection.
\newblock \emph{Electronics}, 13\penalty0 (5):\penalty0 862, 2024.

\bibitem[Kim and Byun(2020)]{LTIR}
Myeongjin Kim and Hyeran Byun.
\newblock Learning texture invariant representation for domain adaptation of semantic segmentation.
\newblock In \emph{IEEE Conf. Comput. Vis. Pattern Recog.}, pages 12975--12984, Seattle, US, 2020.

\bibitem[Lee et~al.(2021)Lee, Rameau, Pan, and Kweon]{3dmotion_seokju}
Seokju Lee, Francois Rameau, Fei Pan, and In~So Kweon.
\newblock Attentive and contrastive learning for joint depth and motion field estimation.
\newblock In \emph{Int. Conf. Comput. Vis.}, pages 4862--4871, Virtual/Online, 2021.

\bibitem[Li et~al.(2021)Li, Gordon, Zhao, Casser, and Angelova]{motion_net}
Hanhan Li, Ariel Gordon, Hang Zhao, Vincent Casser, and Anelia Angelova.
\newblock Unsupervised monocular depth learning in dynamic scenes.
\newblock In \emph{Conf. on Robot Learn.}, pages 1908--1917, London, UK, 2021. PMLR.

\bibitem[Li et~al.(2020)Li, Cao, Wu, and Wong]{li2020generating}
Rui Li, Wenming Cao, Si Wu, and Hau-San Wong.
\newblock Generating target image-label pairs for unsupervised domain adaptation.
\newblock \emph{IEEE Trans. Image Process.}, 29:\penalty0 7997--8011, 2020.

\bibitem[Long et~al.(2015)Long, Shelhamer, and Darrell]{FCN}
Jonathan Long, Evan Shelhamer, and Trevor Darrell.
\newblock Fully convolutional networks for semantic segmentation.
\newblock In \emph{IEEE Conf. Comput. Vis. Pattern Recog.}, pages 3431--3440, Boston, US, 2015.

\bibitem[Lu et~al.(2023)Lu, Li, Song, Xiang, and Hospedales]{lu2023uncertainty}
Zhihe Lu, Da Li, Yi-Zhe Song, Tao Xiang, and Timothy~M. Hospedales.
\newblock Uncertainty-aware source-free domain adaptive semantic segmentation.
\newblock \emph{IEEE Trans. Image Process.}, pages 1--1, 2023.

\bibitem[Mei et~al.(2020)Mei, Zhu, Zou, and Zhang]{IAST}
Ke Mei, Chuang Zhu, Jiaqi Zou, and Shanghang Zhang.
\newblock Instance adaptive self-training for unsupervised domain adaptation.
\newblock In \emph{Eur. Conf. Comput. Vis.}, pages 415--430, Glasgow, UK, 2020. Springer.

\bibitem[Melas-Kyriazi and Manrai(2021)]{PixelMatch}
Luke Melas-Kyriazi and Arjun~K Manrai.
\newblock Pixmatch: Unsupervised domain adaptation via pixelwise consistency training.
\newblock In \emph{IEEE Conf. Comput. Vis. Pattern Recog.}, pages 12435--12445, Virtual/Online, 2021.

\bibitem[Munro and Damen(2020)]{optical_flow_da}
Jonathan Munro and Dima Damen.
\newblock Multi-modal domain adaptation for fine-grained action recognition.
\newblock In \emph{IEEE Conf. Comput. Vis. Pattern Recog.}, pages 122--132, Seattle, US, 2020.

\bibitem[Pan et~al.(2020)Pan, Shin, Rameau, Lee, and Kweon]{IntraDa}
Fei Pan, Inkyu Shin, Francois Rameau, Seokju Lee, and In~So Kweon.
\newblock Unsupervised intra-domain adaptation for semantic segmentation through self-supervision.
\newblock In \emph{IEEE Conf. Comput. Vis. Pattern Recog.}, pages 3764--3773, Seattle, US, 2020.

\bibitem[Pan et~al.(2022)Pan, Hur, Lee, Kim, and Kweon]{fp_3}
Fei Pan, Sungsu Hur, Seokju Lee, Junsik Kim, and In~So Kweon.
\newblock Ml-bpm: Multi-teacher learning with bidirectional photometric mixing for open compound domain adaptation in semantic segmentation.
\newblock In \emph{Eur. Conf. Comput. Vis.}, pages 236--251, Tel Aviv, Israel, 2022. Springer.

\bibitem[Richter et~al.(2016)Richter, Vineet, Roth, and Koltun]{GTA5}
Stephan~R Richter, Vibhav Vineet, Stefan Roth, and Vladlen Koltun.
\newblock Playing for data: Ground truth from computer games.
\newblock In \emph{Eur. Conf. Comput. Vis.}, pages 102--118, Amsterdam, US, 2016. Springer.

\bibitem[Richter et~al.(2017)Richter, Hayder, and Koltun]{VIPER}
Stephan~R Richter, Zeeshan Hayder, and Vladlen Koltun.
\newblock Playing for benchmarks.
\newblock In \emph{Int. Conf. Comput. Vis.}, 2017.

\bibitem[Shin et~al.(2020)Shin, Woo, Pan, and Kweon]{fp_2}
Inkyu Shin, Sanghyun Woo, Fei Pan, and In~So Kweon.
\newblock Two-phase pseudo label densification for self-training based domain adaptation.
\newblock In \emph{Eur. Conf. Comput. Vis.}, pages 532--548, Glasgow, United Kingdom, 2020. Springer.

\bibitem[Tranheden et~al.(2021)Tranheden, Olsson, Pinto, and Svensson]{DACS}
Wilhelm Tranheden, Viktor Olsson, Juliano Pinto, and Lennart Svensson.
\newblock Dacs: Domain adaptation via cross-domain mixed sampling.
\newblock In \emph{Proc. of IEEE/CVF Wint. Conf. on Appli. of Comput. Visi.}, pages 1379--1389, Hawaii, USA., 2021.

\bibitem[Tsai et~al.(2018)Tsai, Hung, Schulter, Sohn, Yang, and Chandraker]{adaptsegnet}
Yi-Hsuan Tsai, Wei-Chih Hung, Samuel Schulter, Kihyuk Sohn, Ming-Hsuan Yang, and Manmohan Chandraker.
\newblock Learning to adapt structured output space for semantic segmentation.
\newblock In \emph{IEEE Conf. Comput. Vis. Pattern Recog.}, pages 7472--7481, Salt Lake City, US, 2018.

\bibitem[Vu et~al.(2019)Vu, Jain, Bucher, Cord, and P{\'e}rez]{advent}
Tuan-Hung Vu, Himalaya Jain, Maxime Bucher, Matthieu Cord, and Patrick P{\'e}rez.
\newblock Advent: Adversarial entropy minimization for domain adaptation in semantic segmentation.
\newblock In \emph{IEEE Conf. Comput. Vis. Pattern Recog.}, pages 2517--2526, Long Beach, US, 2019.

\bibitem[Wang et~al.(2019)Wang, Gao, and Li]{wang2019weakly}
Qi Wang, Junyu Gao, and Xuelong Li.
\newblock Weakly supervised adversarial domain adaptation for semantic segmentation in urban scenes.
\newblock \emph{IEEE Trans. Image Process.}, 28\penalty0 (9):\penalty0 4376--4386, 2019.

\bibitem[Wang et~al.(2022)Wang, Zhao, Zhang, Guo, Zang, Gu, Li, and Jiao]{wang2022cluster}
Shuang Wang, Dong Zhao, Chi Zhang, Yuwei Guo, Qi Zang, Yu Gu, Yi Li, and Licheng Jiao.
\newblock Cluster alignment with target knowledge mining for unsupervised domain adaptation semantic segmentation.
\newblock \emph{IEEE Trans. Image Process.}, 31:\penalty0 7403--7418, 2022.

\bibitem[Wang et~al.(2020)Wang, Yu, Wei, Feris, Xiong, Hwu, Huang, and Shi]{SIM}
Zhonghao Wang, Mo Yu, Yunchao Wei, Rogerio Feris, Jinjun Xiong, Wen-mei Hwu, Thomas~S Huang, and Honghui Shi.
\newblock Differential treatment for stuff and things: A simple unsupervised domain adaptation method for semantic segmentation.
\newblock In \emph{IEEE Conf. Comput. Vis. Pattern Recog.}, pages 12635--12644, Seattle, US, 2020.

\bibitem[Wu et~al.(2005)Wu, Otoo, and Shoshani]{connected_component}
Kesheng Wu, Ekow Otoo, and Arie Shoshani.
\newblock Optimizing connected component labeling algorithms.
\newblock In \emph{Medic. Imagi.: Image Processing}. SPIE, 2005.

\bibitem[Xie et~al.(2023)Xie, Li, Li, Liu, Huang, and Wang]{SePiCo}
Binhui Xie, Shuang Li, Mingjia Li, Chi~Harold Liu, Gao Huang, and Guoren Wang.
\newblock Sepico: Semantic-guided pixel contrast for domain adaptive semantic segmentation.
\newblock \emph{IEEE Trans. Pattern Anal. Mach. Intell.}, 2023.

\bibitem[Xie et~al.(2021)Xie, Wang, Yu, Anandkumar, Alvarez, and Luo]{segformer}
Enze Xie, Wenhai Wang, Zhiding Yu, Anima Anandkumar, Jose~M Alvarez, and Ping Luo.
\newblock Segformer: Simple and efficient design for semantic segmentation with transformers.
\newblock \emph{Adv. Neural Inform. Process. Syst.}, 34:\penalty0 12077--12090, 2021.

\bibitem[Xing et~al.(2022)Xing, Guan, Huang, and Lu]{TPS}
Yun Xing, Dayan Guan, Jiaxing Huang, and Shijian Lu.
\newblock Domain adaptive video segmentation via temporal pseudo supervision.
\newblock In \emph{Eur. Conf. Comput. Vis.}, pages 621--639, Tel Aviv, Israel, 2022. Springer.

\bibitem[Yuan et~al.(2022)Yuan, Zhao, Shao, Yuan, and Wang]{yuan2022birds}
Bo Yuan, Danpei Zhao, Shuai Shao, Zehuan Yuan, and Changhu Wang.
\newblock Birds of a feather flock together: Category-divergence guidance for domain adaptive segmentation.
\newblock \emph{IEEE Trans. Image Process.}, 31:\penalty0 2878--2892, 2022.

\bibitem[Zhang et~al.(2021)Zhang, Zhang, Zhang, Chen, Wang, and Wen]{ProDA}
Pan Zhang, Bo Zhang, Ting Zhang, Dong Chen, Yong Wang, and Fang Wen.
\newblock Prototypical pseudo label denoising and target structure learning for domain adaptive semantic segmentation.
\newblock In \emph{IEEE Conf. Comput. Vis. Pattern Recog.}, pages 12414--12424, Virtual/Online, 2021.

\bibitem[Zhang et~al.(2019)Zhang, Zhang, Liu, and Tao]{CAG_UDA}
Qiming Zhang, Jing Zhang, Wei Liu, and Dacheng Tao.
\newblock Category anchor-guided unsupervised domain adaptation for semantic segmentation.
\newblock \emph{Adv. Neural Inform. Process. Syst.}, 32, 2019.

\bibitem[Zhao et~al.(2022)Zhao, Zhong, Luo, Lee, and Sebe]{zhao2022source}
Yuyang Zhao, Zhun Zhong, Zhiming Luo, Gim~Hee Lee, and Nicu Sebe.
\newblock Source-free open compound domain adaptation in semantic segmentation.
\newblock \emph{IEEE Trans. Circuit Syst. Video Technol.}, 32\penalty0 (10):\penalty0 7019--7032, 2022.

\bibitem[Zheng and Yang(2019)]{MRNet}
Zhedong Zheng and Yi Yang.
\newblock Unsupervised scene adaptation with memory regularization in vivo.
\newblock 2019.

\bibitem[Zheng and Yang(2021)]{rectifying}
Zhedong Zheng and Yi Yang.
\newblock Rectifying pseudo label learning via uncertainty estimation for domain adaptive semantic segmentation.
\newblock \emph{Int. J. Comput. Vis.}, 129\penalty0 (4):\penalty0 1106--1120, 2021.

\bibitem[Zhou et~al.(2021)Zhou, Wang, Chu, Yang, Bai, and Xu]{zhou2020affinity}
Wei Zhou, Yukang Wang, Jiajia Chu, Jiehua Yang, Xiang Bai, and Yongchao Xu.
\newblock Affinity space adaptation for semantic segmentation across domains.
\newblock \emph{IEEE Trans. Image Process.}, 30:\penalty0 2549--2561, 2021.

\bibitem[Zou et~al.(2018)Zou, Yu, Kumar, and Wang]{CBST}
Yang Zou, Zhiding Yu, BVK Kumar, and Jinsong Wang.
\newblock Unsupervised domain adaptation for semantic segmentation via class-balanced self-training.
\newblock In \emph{Eur. Conf. Comput. Vis.}, pages 289--305, Munich, Germany, 2018.

\bibitem[Zou et~al.(2019)Zou, Yu, Liu, Kumar, and Wang]{CRST}
Yang Zou, Zhiding Yu, Xiaofeng Liu, BVK Kumar, and Jinsong Wang.
\newblock Confidence regularized self-training.
\newblock In \emph{Int. Conf. Comput. Vis.}, pages 5982--5991, Seoul, South Korea, 2019.

\end{thebibliography}
}


\end{document}